\documentclass{article}

\usepackage{microtype}
\usepackage{graphicx}
\usepackage{subfigure}
\usepackage{amsmath}
\usepackage{tablefootnote}

\usepackage{multirow}
\usepackage[table,xcdraw]{xcolor}
\usepackage{tabularx}
\usepackage{booktabs} %
\usepackage{xspace}
\usepackage{caption}
\usepackage{circledsteps}
\usepackage{subcaption}
\usepackage{enumitem}
\usepackage{hyperref}

\newcommand{\etespeedupfp}{$7.73 \times$\xspace} 
\newcommand{\etespeedupsq}{$2.53 \times$\xspace} 

\newcommand{\qq}{Atom} %

\newcommand{\kv}{KV-cache}
\newcommand{\woq}{weight-only quantization}
\newcommand{\waq}{weight-activation quantization}
\newcommand{\lm}{Llama}

\usepackage[accepted]{mlsys2024}

\mlsystitlerunning{\qq}

\begin{document}

\twocolumn[
\mlsystitle{\qq: Low-Bit Quantization for Efficient and Accurate LLM Serving}
\mlsyssetsymbol{intern}{*}

\begin{mlsysauthorlist}
\mlsysauthor{Yilong Zhao}{sjtu,uw,intern}
\mlsysauthor{Chien-Yu Lin}{uw}
\mlsysauthor{Kan Zhu}{uw}
\mlsysauthor{Zihao Ye}{uw}
\mlsysauthor{Lequn Chen}{uw}
\mlsysauthor{Size Zheng}{uw,pku,intern}\\
\mlsysauthor{Luis Ceze}{uw,octoai}
\mlsysauthor{Arvind Krishnamurthy}{uw}
\mlsysauthor{Tianqi Chen}{octoai,cmu}
\mlsysauthor{Baris Kasikci}{uw}
\end{mlsysauthorlist}

\mlsysaffiliation{sjtu}{Department of Computer Science and Engineering, Shanghai Jiao Tong University, Shanghai, China}
\mlsysaffiliation{uw}{School of Computer Science \& Engineering, University of Washington, Seattle, United States}
\mlsysaffiliation{cmu}{School of Computer Science, Carnegie Mellon University, Pittsburgh, United States}
\mlsysaffiliation{pku}{School of Computer Science, Peking University, Beijing, China}
\mlsysaffiliation{octoai}{OctoAI}

\mlsyscorrespondingauthor{Yilong Zhao}{zhaoyilong217@sjtu.edu.cn}

\mlsyskeywords{Machine Learning System, Quantization, Efficient AI}

\vskip 0.3in

\begin{abstract}
The growing demand for Large Language Models (LLMs) in applications such as content generation, intelligent chatbots, and sentiment analysis poses considerable challenges for LLM service providers. To efficiently use GPU resources and boost throughput, batching multiple requests has emerged as a popular paradigm;  to further speed up batching, LLM quantization techniques reduce memory consumption and increase computing capacity. However, prevalent quantization schemes (e.g., 8-bit weight-activation quantization) cannot fully leverage the capabilities of modern GPUs, such as 4-bit integer operators, resulting in sub-optimal performance. 

To maximize LLMs' serving throughput, we introduce \qq{}, a low-bit quantization method that achieves high throughput improvements with negligible accuracy loss. \qq{} significantly boosts serving throughput by using low-bit operators and considerably reduces memory consumption via low-bit quantization. It attains high accuracy by applying a novel mixed-precision and fine-grained quantization process. We evaluate \qq{} on 4-bit \waq{} in the serving context. \qq{} improves end-to-end throughput (token/s) by up to \etespeedupfp{} compared to the FP16 and  by \etespeedupsq{} compared to INT8 quantization, while maintaining the same latency target.

\end{abstract}
]

\printAffiliationsAndNotice{\mlsysIntern}

\section{Introduction}

Large Language Models (LLMs) are increasingly being integrated into our work routines and daily lives, where we use them for summarization, code completion, and decision-making. Studies report that ChatGPT has over 100 million users, with more than 1 billion website accesses per month~\cite{servenumber}. Furthermore, the size and capabilities of LLMs continue to grow to accommodate a broader range of tasks. The high inference demand and model complexity have significantly increased the operational costs, i.e., compute/memory and energy, for LLM service providers to near \$1 million daily~\cite{servemoney}.

Unsurprisingly, optimizing LLM serving is becoming a pressing concern. Most efforts have focused on improving LLM serving throughput, which is typically achieved by batching requests from various users~\cite{orca,batchblog,vllm}. Batching multiple requests increases compute intensity and amortizes the cost of loading weight matrices, thereby improving throughput. Prior work has explored LLM quantization techniques to further improve batching efficiency. These techniques employ smaller data types to replace 16-bit floating point (FP16) values, thereby reducing memory consumption and accelerating computation~\cite{awq,smoothquant}.

However, current quantization schemes do not leverage the full extent of capabilities provided by emerging efficient low-bit hardware support (e.g., Nvidia Ampere~\cite{ampere} and Qualcomm Hexagon~\cite{Snapdragon}). For instance, several prior approaches have explored \woq{}~\cite{awq,gptq}. In these quantization schemes, weights are quantized to a low-bit representation (e.g., INT3), whereas activations remain in a floating point representation (e.g., FP16). Consequently, weights must be dequantized to the appropriate floating point representation (e.g., FP16) before being multiplied with activations using floating point representation. Therefore, even though weight-only quantization reduces memory consumption, it still requires costly floating-point arithmetic, which is inefficient, especially for large batch sizes.

\begin{figure}
    \centering
    \includegraphics[width=0.5\columnwidth]{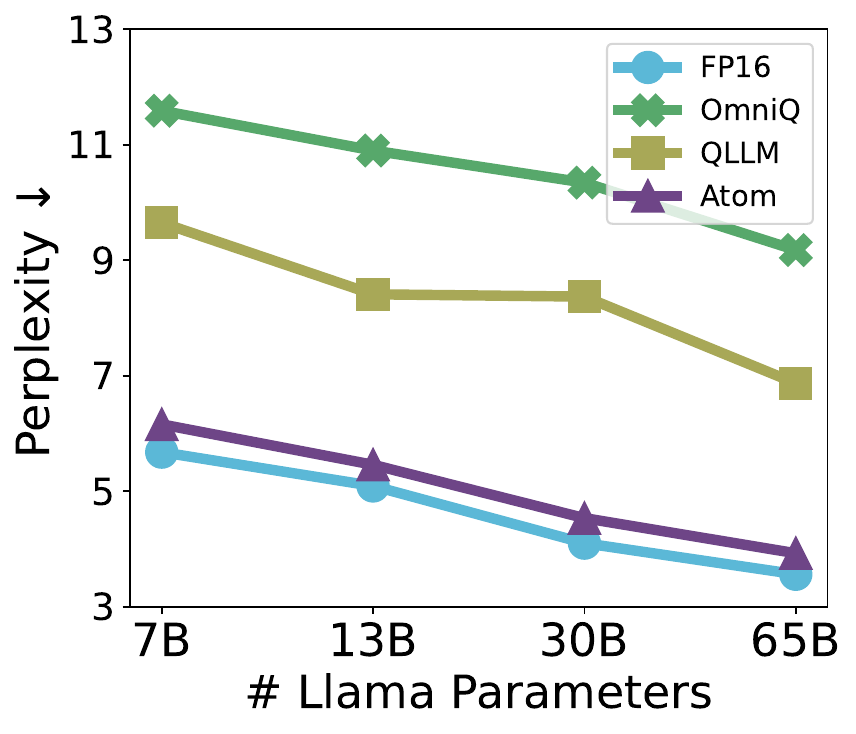}
    \vspace{-5pt}
    \caption{WikiText2 perplexity on \lm{} models with different 4-bit \waq{} mechanisms. Atom maintains perplexity results close to the FP16 baseline across all model sizes.}
    \label{fig:intro-performance-ppl}
    \vspace{-0.2in}
\end{figure}

Another prominent quantization scheme is \waq{}, where both weights and activations are quantized to low-bit representations. In this scheme, weights and activations can be directly multiplied using low-precision arithmetic units. This quantization approach has greater potential to achieve higher inference throughput than \woq{} due to the efficient low-bit hardware support. For example, A100 GPUs can reach $1248$ TOPS of INT4 and $624$ TOPS of INT8 as opposed to only $312$ TFLOPS for FP16 with Tensor Cores~\cite{a100-spec}. Prior works such as LLM.INT8()~\cite{llmint8} and SmoothQuant~\cite{smoothquant} explored INT8 \waq{} and achieved near no accuracy loss. However, INT8 quantization still cannot utilize lower bit arithmetic such as INT4 Tensor Cores~\cite{tc-spec}. In addition, INT8 quantization remains sub-optimal for reducing the large memory consumption in LLM serving, where both model parameters and batched \kv{} consume large memory~\cite{flexgen,zhang2023h2o}. For lower-bit \waq{}, recent works such as OmniQuant~\cite{shao2023omniquant} and QLLM~\cite{liu2023qllm} have proposed to quantize LLMs down to 4-bit. However, their techniques still show a significant perplexity increase compared to the FP16 baseline as shown in Figure~\ref{fig:intro-performance-ppl}. Therefore, determining how to accurately quantize LLMs into low-bit representations while maintaining hardware efficiency remains an open area of research.

\begin{figure}[t!]
\centering
\includegraphics[width=0.9\columnwidth]{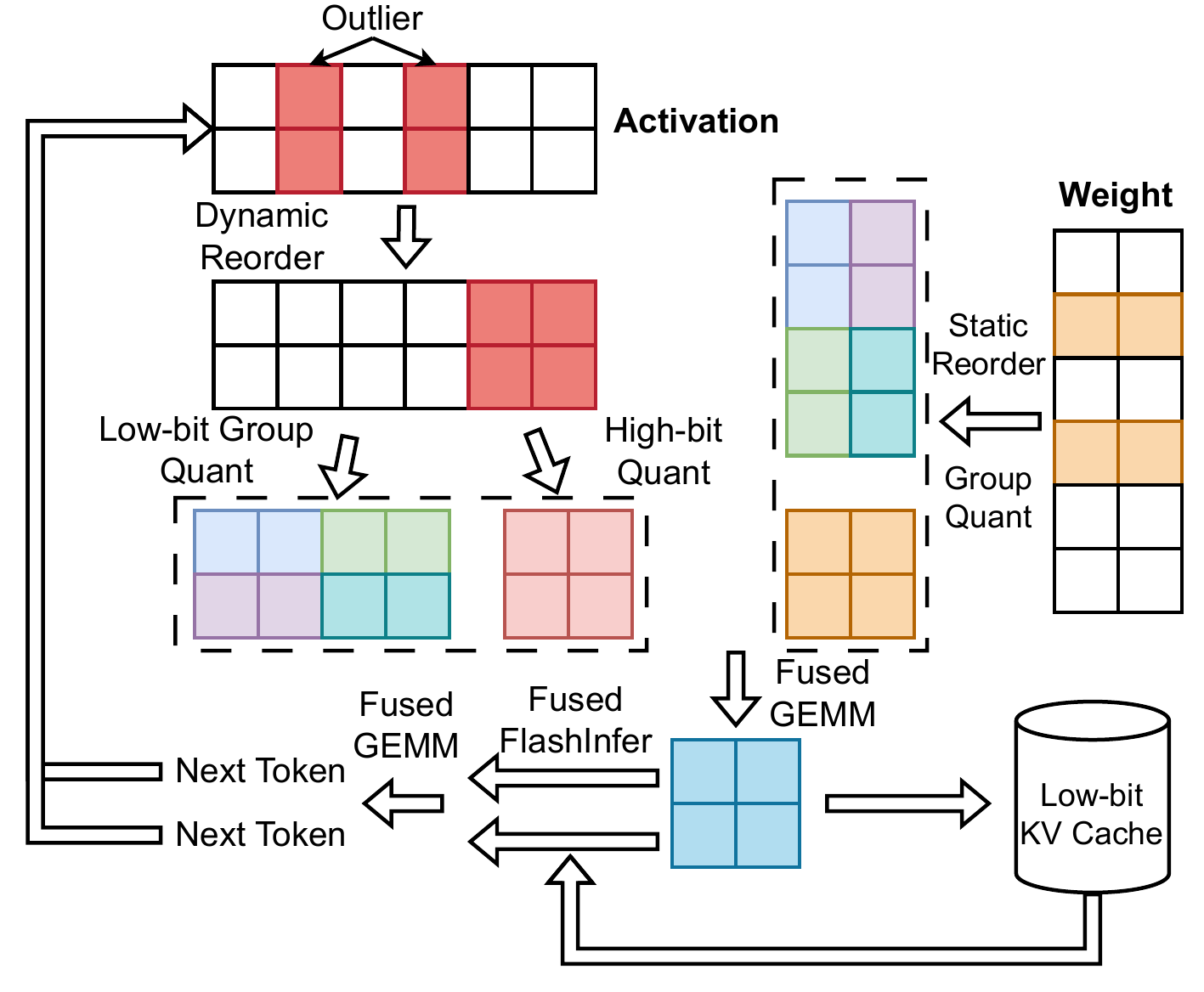}
\vspace{-5pt}
\caption{Overview of \qq{}'s design. For activation matrices, we dynamically reorder the channels to pick out the outliers. Then, we apply low-bit group quantization to the normal values while using high-bit precision for outliers. For weight matrices, the quantization process can be done statically. We perform fused GEMM and fused FlashInfer~\cite{flashinfer} to boost throughput. We also adopt a quantized \kv{} to reduce memory movement.}
\vspace{-1.5em}
\label{fig:design}
\end{figure}

In this work, we introduce \qq{}, an accurate low-bit \waq{} for LLMs that efficiently use modern hardware. To maintain accuracy, \qq{} incorporates three key quantization designs: (1) It adopts mixed-precision quantization, which retains a small but salient number of activations and weights in high precision to preserve accuracy. (2) It employs fine-grained group quantization on both weights and activations, which naturally reduces quantization errors. (3) Instead of pre-calculating quantization parameters for activations, \qq{} dynamically quantizes activations to best capture the distribution of each input.

Although these quantization optimizations can improve quantization accuracy, they may not utilize the underlying hardware efficiently without a bespoke design. For example, the mixed-precision technique could lead to irregular memory accesses and performance slowdown~\cite{oliveguo}; matrix multiplications with group quantization are not well-supported in kernel libraries; and dynamic quantization of activations incurs extra computation~\cite{smoothquant}. To ensure high hardware efficiency and minimize quantization overheads, \qq{}: (1) reorders activations and weights to maintain regular memory accesses for mixed-precision operations, (2) fuses quantization and reordering operations into existing operators to mitigate the overheads, (3) further quantizes outliers into 8-bit to keep a balance between accuracy and efficiency and (4) quantizes the \kv{} into low-bit representations to reduce memory movement. We illustrate \qq{}'s quantization workflow in Figure~\ref{fig:design}.

To validate \qq{}'s feasibility, we integrate it into an end-to-end serving framework~\cite{punica}. For our special matrix multiplications with mixed-precision and group quantization, we implement customized CUDA kernels that utilize low-bit tensor cores. Experiments on popular datasets show that \qq{} has negligible accuracy loss ($1.4$\% average zero-shot accuracy drop, $0.3$ WikiText2 perplexity increase for \lm{}-65B) when quantizing models to 4-bit (for both weights and activations), while prior works suffer larger accuracy loss under the same precision (see Table~\ref{table:llama_zeroshot}).

When comparing end-to-end serving throughput to different precisions and quantization schemes, \qq{} improves throughput by up to $7.7\times$, $5.5\times$, and $2.5\times$ relative to FP16, W4A16, and W8A8, respectively, while achieving similar latency (see Figure~\ref{fig:e2e-all}). These results show that \qq{} can accurately quantize LLMs into low-bit precision while achieving high serving throughput.

In summary, we contribute the following:
\begin{itemize}[noitemsep,topsep=0pt,parsep=0pt,partopsep=0pt]
    \item  A comprehensive performance analysis of LLM serving workloads that pinpoints the efficiency benefit of low-bit \waq{}.
    \item \qq{}, an accurate low-bit \waq{} algorithm that combines (1) mixed-precision with channel reordering, (2) fine-grained group quantization, (3) dynamic activation quantization to minimize quantization errors, and (4) \kv{} quantization.
    \item   An integrated LLM serving framework for which we codesign an efficient inference workflow, implement low-bit GPU kernels and demonstrate practical end-to-end throughput and latency of \qq{}. 
    \item A comprehensive evaluation of  \qq{}, which shows that it improves LLM serving throughput by up to $7.7\times$ with only a slight accuracy loss.
\end{itemize}
\section{Background}
\label{sec:background}
Quantization techniques use discrete low-bit values to approximate high-precision floating points. Since integers represent a uniform range, quantizing floating point values into integers is widespread due to simplicity and hardware efficiency~\cite{integeronly,deepcompression}. Typical quantization involves two steps: determining the quantization parameters (which consist of scale and zero point) and calculating the quantized tensor. For uniform asymmetric quantization, the scale $s$ and zero point $z$ are determined by~\cite{quantizewhite}:

\begin{equation}
    s = \frac{\max(X) - \min(X)}{2^n -1} \cdot c, z =\lfloor \frac{-\min(X)}{s}\rceil,
\end{equation}

where $X$ is the input tensor, $n$ is the quantization bit-width, and $c$ is the clipping factor used to reduce the dynamic range of quantization to mitigate the effect of outlier values. The elements in quantized tensor can be calculated by:

$$
    \Bar{X} = \text{clamp} (\lfloor \frac{X}{s} \rceil + z, 0, 2^n -1).
$$

We can further simplify this equation for symmetric quantization:
$$
    s = \frac{2 \cdot \max(|X|)}{2^n -1} \cdot c 
    $$
    $$
    \Bar{X} = \text{clamp} (\lfloor \frac{X}{s} \rceil, -2^{n-1}, 2^{n-1} -1).
$$

Quantization parameters $s$ and $z$ can be calculated either statically using calibration data or dynamically during inference time with runtime statistics. Thus, quantization approaches can be classified as \textit{static} or \textit{dynamic}.

For LLMs, we can apply quantization on both activation and weight matrices (\waq{}) or just the latter (\woq{}). However, asymmetric \waq{} can lead to additional calculations during matrix multiplication since:
$$
    W\cdot X = s_W(\Bar{W} -z_W) \cdot s_x(\Bar{X} - z_x),
$$
where three additional cross-terms need to be calculated for using low-bit arithmetic units. Therefore, we apply symmetric quantization in this work for efficiency.

Different trade-offs between accuracy and efficiency can be achieved by quantization with different granularity: For \textbf{per-tensor} quantization, all the values in the tensor share one set of scale and zero-point~\cite{quantizewhite}. For \textbf{per-channel (token)} quantization, we calculate scale and zero-point for a row or a column of the tensor~\cite{smoothquant}. We denote the channel as the last dimension of the input matrix. Each channel can be further divided into several sub-groups, and quantization is individually performed on each group, which is called \textbf{per-group} quantization~\cite{awq}. The finer the granularity, the more precise the quantization, but the higher the overhead. In this work, we adopt group quantization for higher accuracy with dedicated kernels to manage the overhead, as shown in \S~\ref{sec:group}.

\section{Performance analysis of low-bit LLM serving}
\label{sec:analyze}
In this section, we first analyze the performance bottleneck of LLM inference in serving scenarios and then establish the importance of low-bit \waq{}. 

\begin{figure}[bt!]
    \centering
    \includegraphics[width=0.80\linewidth]{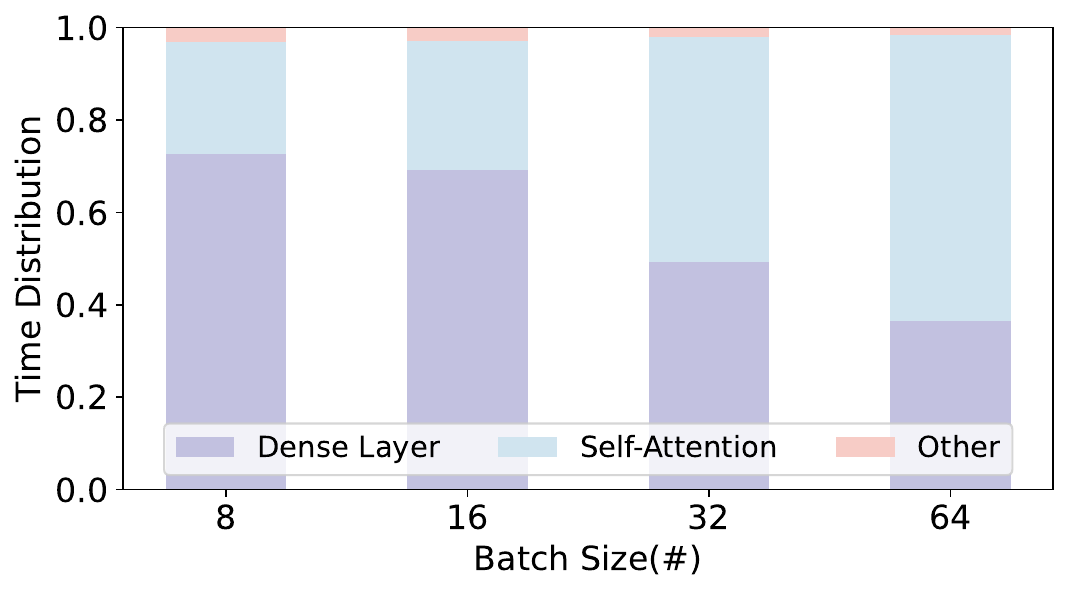}
    \vspace{-5pt}
    \caption{
        Runtime breakdown of \lm{}-7b inference with different batch sizes. The dense layer represents the batched K, Q, V generation, O projection, and MLP. The self-attention layer is implemented by FlashInfer~\cite{flashinfer} integrated with PageAttention~\cite{vllm}. Results indicate that the dense and self-attention layers together account for over $90$\% of the execution time, thereby constraining the throughput.
    }
    \label{fig:breakdown}
    \vspace{-0.1in}
\end{figure}

Due to high demand, LLM serving is throughput-oriented. However, the auto-regressive decode stage of LLM inference only takes one token as input and generates the next token, thus relying on matrix-vector multiplication (GEMV)~\cite{taming}. Since GEMV needs to load a large weight matrix while only performing a few multiplications, it is heavily memory-bound. It thus causes GPU under-utilization, which results in low compute intensity (computation-to-IO ratio) and, thereby, low throughput~\cite{williams2009roofline}. To mitigate this problem, batching is widely used by combining the input from multiple requests to perform dense layer (K,Q,V generation, O projection, and MLP) matrix multiplications and increase compute intensity, therefore GPU utilization~\cite{scaleinference,orca,punica,distserve}.

To further exploit the batching effect and boost throughput, the input matrices of the dense layer of the decode and prefill stages are batched together to form larger matrices~\cite{splitwise}. Given large batch sizes, the dense layer ends up having compute-bound matrix-matrix multiplications (GEMM). However, though self-attention layers in the decode stage are also GEMV operations, they cannot benefit from batching. Since different inference requests do not share the \kv{} with different context histories, cross-request data cannot be batched for reuse, resulting in no efficiency benefit. Even with several optimizations such as FlashAttention~\cite{dao2023flashattention2} or Group Query Attention~\cite{ainslie2023gqa}, the self-attention layers are still bounded by the large memory movement of \kv{}.

After applying the batching technique, we measure the time breakdown of different operators under different batch sizes. As Figure~\ref{fig:breakdown} shows, both the dense and self-attention layers act as bottlenecks to throughput, consuming over $90$\% of the processing time. Consequently, we employ quantization mechanisms to expedite both dense and self-attention layers. 

\begin{figure}[bt!]
    \centering
    \subfigure[Weight-activation quantization]{
        \includegraphics[width=0.47\columnwidth]{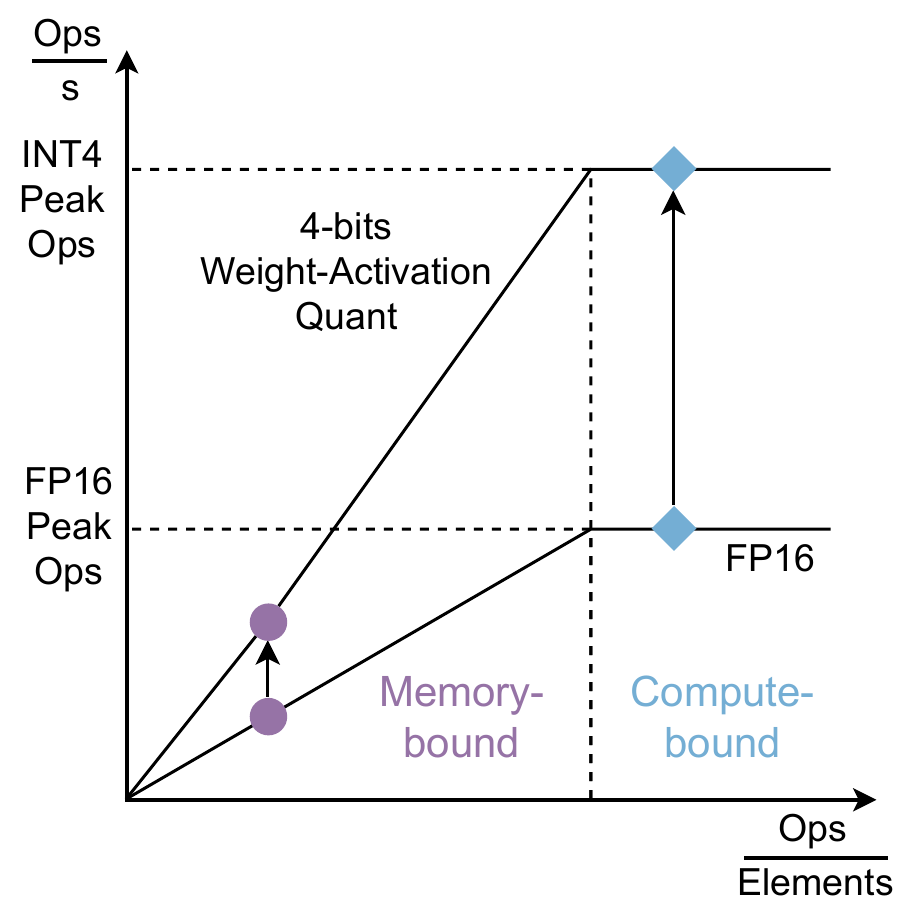}
        \label{fig:roofline-waq}
    }
    \subfigure[Weight-only quantization]{
        \includegraphics[width=0.47\columnwidth]{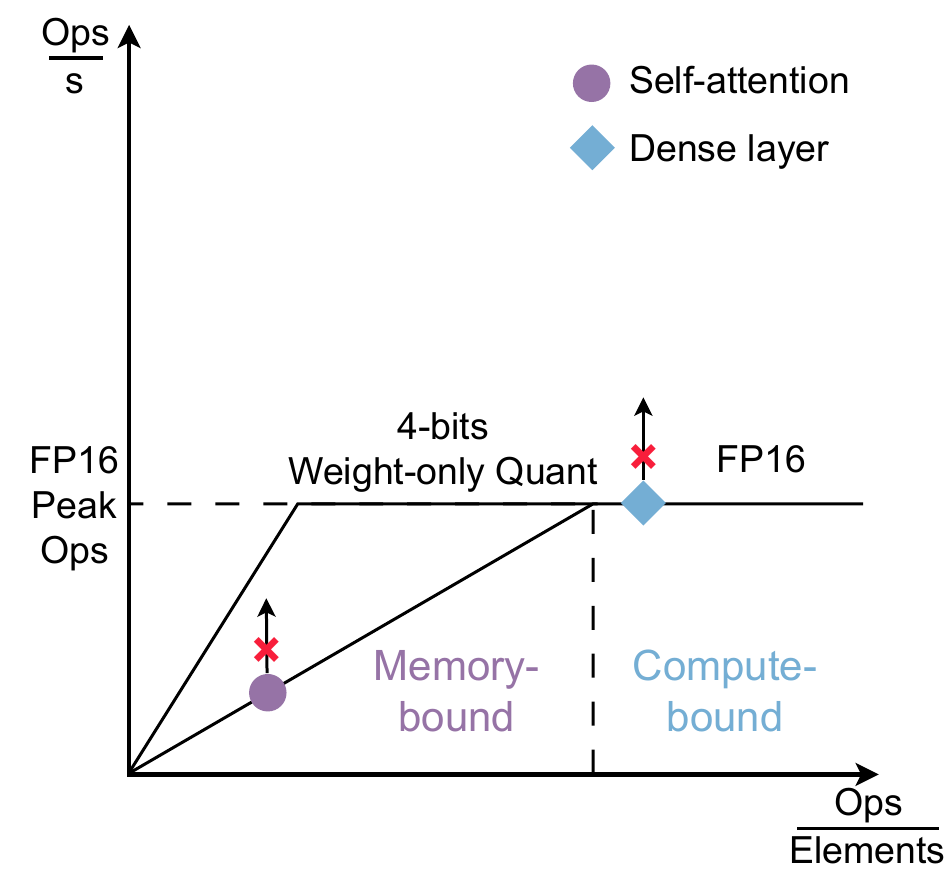}
        \label{fig:roofline-woq}
    }
    \vspace{-5pt}
    \caption{A roofline model of different quantization approaches that characterizes operators by their arithmetic intensity, which is defined as $\text{Ops}/\text{Elements}$. At large batch sizes, the dense layer is compute-bound, which has a large arithmetic intensity, whereas self-attention consistently exhibits a lower arithmetic intensity.}
    \label{fig:roofline}
    \vspace{-0.1in}
\end{figure}

We use the Roofline model~\cite{williams2009roofline} to evaluate the effect of different quantization approaches in serving scenarios. As Figure~\ref{fig:roofline-waq} shows, \waq{} has higher dense layer compute throughput due to the efficient low-bit hardware arithmetic. It also increases the throughput of the self-attention layer by reducing the size of the \kv{}, thus decreasing memory movement. However, as Figure~\ref{fig:roofline-woq} shows, \woq{} fails to improve dense layer throughput since dequantization must be performed before matrix multiplications, yielding calculations still in the floating point format. On the other hand, \woq{} fails to quantize the \kv{}, yielding no benefit for self-attention layers. We further quantify the effect of different quantization techniques in Figures~\ref{fig:batch-dense} and \ref{fig:batch-self-attn} in \S\ref{sec:evaluation} with kernel profiling.

In summary, the low-bit \waq{} is superior to \woq{} in terms of enhancing the throughput in the serving scenario because it accelerates both the dense and self-attention layers. In the following sections, we demonstrate how \qq{} delivers high throughput while still maintaining high accuracy with the low-bit \waq{}.

\section{Design}

\begin{figure}[t!]
    \centering
    \subfigure[Activation mean values per channel.]{
        \includegraphics[width=0.45\columnwidth]{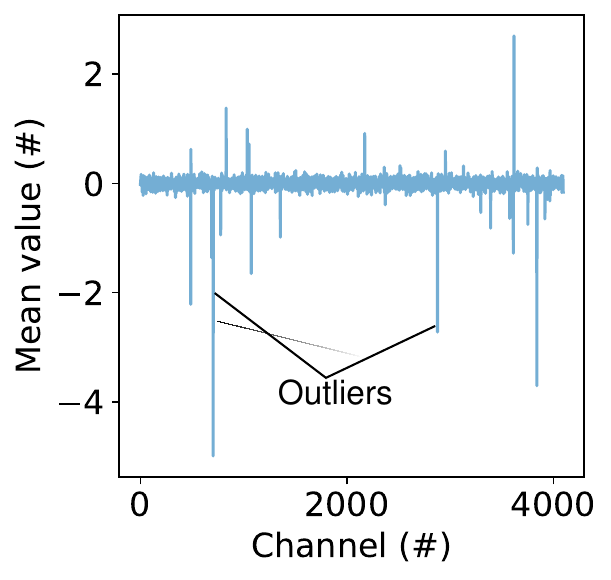}
        \label{fig:design-outlier}
    }
    \hspace{0.005\columnwidth}
    \subfigure[Mean values after reordering.]{
        \includegraphics[width=0.45\columnwidth]{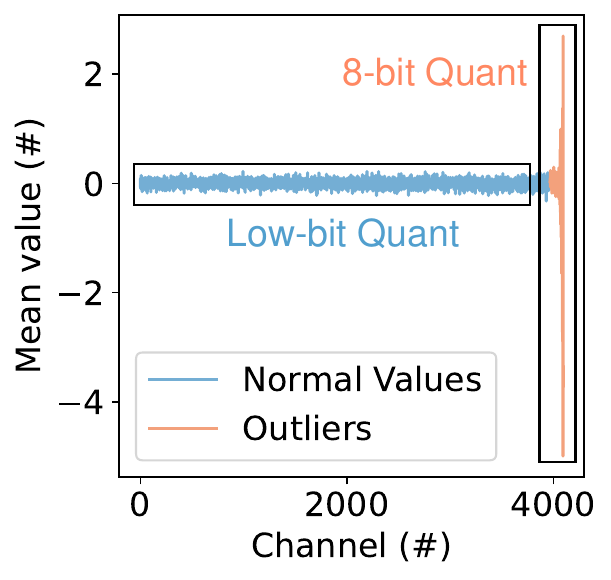}
        \label{fig:design-mixed-precision}
    }
    \vspace{-4pt}
    \caption{
        Sampled value of an activation matrix from \lm{}-7b. (a) The activation matrix contains outlier channels, which result in large quantization errors. (b) \qq{} reorders these outlier channels to the end of the matrix and uses higher precision to quantize them while keeping regular memory access.
    }
    \label{fig:design-profile}
    \vspace{-1em}
\end{figure}

Low-bit precision enables efficient utilization of the underlying hardware, leading to increased throughput. However, it is challenging to maintain high accuracy with a low-bit representation. To quantize LLMs to extremely low-bit precision while keeping accuracy, we incorporate a suite of quantization mechanisms tailored to LLM characteristics. These mechanisms include mixed-precision quantization with channel reordering, fine-grained group quantization, and dynamic quantization. We demonstrate the accuracy gain thanks to these techniques with ablation study in Table~\ref{table:quant_ablation}. \qq{} also applies low-bit quantization on \kv{}, which further boosts the efficiency. The subsequent subsections delve into the specifics of each mechanism and its advantages, followed by a detailed description of the end-to-end workflow.

\begin{figure*}[t!]
\centering
\includegraphics[width=\textwidth]{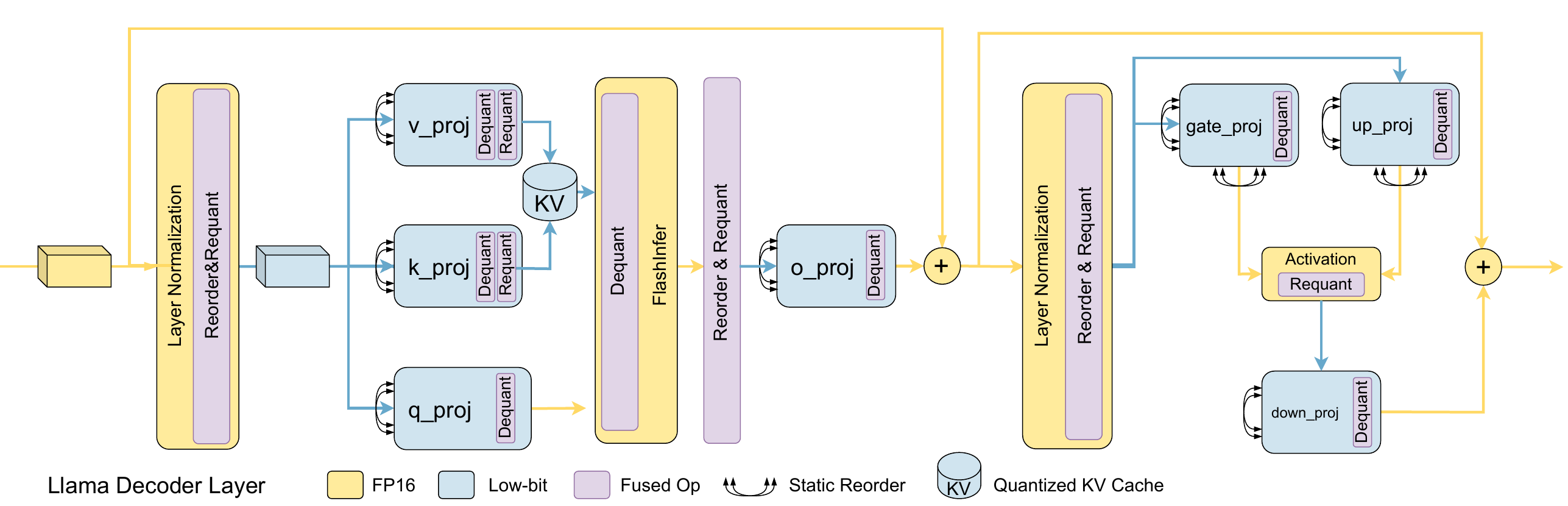}
\caption{Overview of \qq{} workflow on \lm{} model family. \qq{} carefully manages the overhead of quantization operators by fusing them into existing operators. For the compute-bound operators, \qq{} utilizes efficient low-bit hardware support. For the memory-bound self-attention layer, \qq{} quantizes \kv{} to further enhance the throughput. We implement dedicated kernels for each fused operator.}
\vspace{-0.15in}
\label{fig:workflow}
\end{figure*}

\subsection{Mixed-precision quantization}
Prior works observed that a key challenge of LLM quantization is the outlier phenomena in activations~\cite{llmint8,awq}. As Figure~\ref{fig:design-outlier} shows, a few channels exhibit large magnitudes that are several orders greater than those of other channels, which are called outliers. The large dynamic range of these outliers can substantially increase the quantization error. Therefore, efficiently handling the outliers is crucial in low-bit quantization.

One intuitive way to effectively mitigate this challenge is to quantize outliers and normal values separately, into low and high bits, which is referred to as a \textit{mixed-precision} method. As Figure~\ref{fig:design-mixed-precision} shows, after we remove the outliers, the remaining channels are much more uniform, which can be effectively expressed by low-bit values. Our results indicate that 8-bit representations, such as FP8~\cite{micikevicius2022fp8} and INT8, are sufficient to express outliers (See Table~\ref{table:quant_ablation}). Since INT8 is widely supported by hardware implementations (e.g., NVIDIA Tensor Core~\cite{ampere}), \qq{} applies INT8 quantization for outliers.

\begin{figure}[t!]
\centering
\includegraphics[width=0.95\columnwidth]{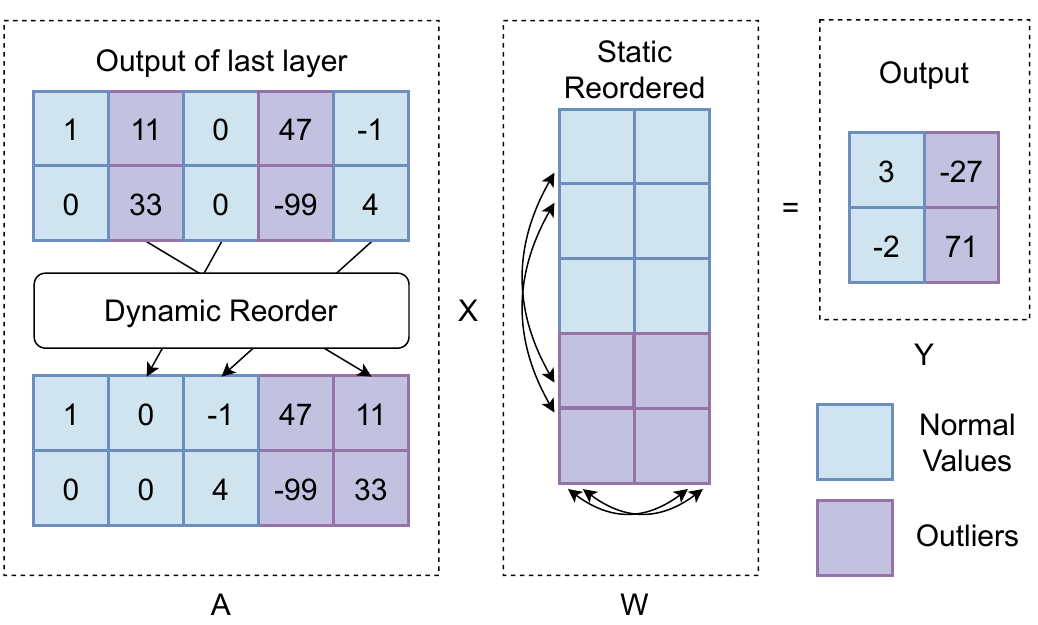}
\vspace{-5pt}
\caption{\qq{} dynamically reorders activation (A) to move the outlier channels to the end of the matrix, with the reorder indices determined in offline calibration. The weight matrix (W) is statically reordered to remain aligned with the corresponding activation channels, which guarantees the correctness of the output result.}
\label{fig:design-reorder}
\vspace{-0.25in}
\end{figure}

The primary concern with mixed-precision quantization is its irregular memory accesses~\cite{llmint8,oliveguo}, which leads to poor hardware efficiency. To apply mixed-precision quantization while maintaining regular memory access, \qq{} re-purposes the reordering technique introduced in RPTQ~\cite{yuan2023rptq}, where the objective is to improve quantization accuracy. As Figure~\ref{fig:design-reorder} shows, \qq{} reorders the scattered outlier channels of activations to the end of the matrix, which enables the efficient implementation of mixed-precision. To guarantee the equivalence of the computation result, the weight matrices need to be reordered with the corresponding reorder indices of activations. Since the outlier channels can be identified offline using calibration data~\cite{llmint8}, the reordering of weight matrices incurs a one-time cost. However, the reordering of activation matrices still needs to be performed online, which can be expensive. To mitigate this, \qq{} fuses the activation matrix reordering operators into prior operators, which significantly reduces the reordering overhead to less than $0.5$\% of runtime.

\subsection{Fine-grained group quantization}
\label{sec:group}
Even if \qq{} quantizes outliers and normal values separately, the latter is still challenging to perform accurately due to the limited representation capability of 4-bit precision~(Section~\ref{sec:ablation}). To further enhance accuracy, \textit{group quantization} is widely adopted~\cite{awq,quantizewhite}, which divides the matrix into subgroups and performs quantization within each subgroup. For example, a group size of $128$ implies that every contiguous sequence of $128$ elements is treated as a single group, which is quantized independently.

\begin{figure}[tb!]
\centering
\includegraphics[width=0.9\columnwidth]{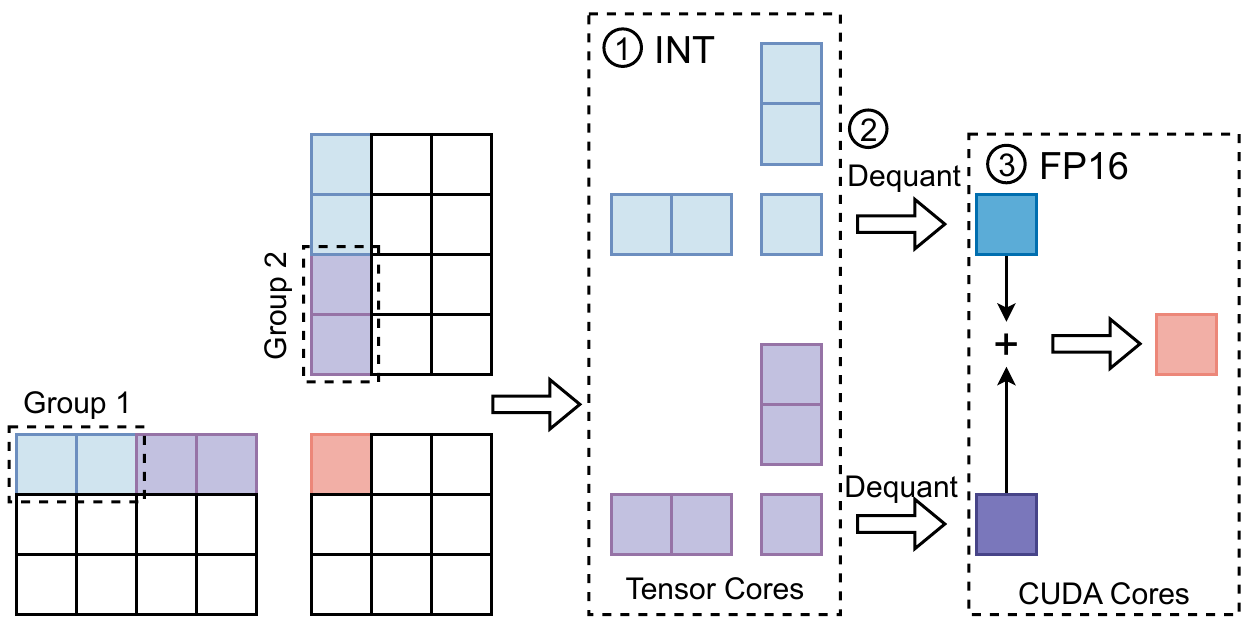}
\vspace{-5pt}
\caption{Overview of the fused GEMM operator. The multiplication of each group is first computed by units with efficient low-bit support, i.e., Tensor Cores (Step \Circled{1}). The result is then dequantized and subsequently accumulated with typical FP16 units (Step \Circled{2}, \Circled{3}). Note that all operations are fused in a single pipeline.}
\label{fig:matmul-design}
\vspace{-0.225in}
\end{figure}

Group quantization offers a trade-off between accuracy improvements and dequantization overheads, especially in \waq{}. Prior works have not investigated how to efficiently incorporate group dequantization into the delicate GEMM pipeline, i.e., MMA pipeline~\cite{Thakkar_CUTLASS_2023}. \qq{} proposes a fusion technique as shown in Figure~\ref{fig:matmul-design}, which contributes to an efficient GEMM kernel with practical speedup (See \S\ref{sec:matrix}). \qq{} first calculates the matrix multiplication of the activation groups with the corresponding weight groups and obtains temporary results using efficient low-bit hardware, i.e. Tensor Cores (Step \Circled{1}). \qq{} then adds multiple temporary results together to get the GEMM result. However, since \qq{} performs fine-grained quantization for each activation and weight group, each temporary result has different quantization parameters. Therefore, \qq{} first dequantizes all temporary results to the FP16 representation with CUDA Cores (Step \Circled{2}) and then performs addition (Step \Circled{3}). To manage the overhead, we fuse dequantization and summation into the GEMM kernel, to be specific, into the MMA pipeline. Therefore, the additional operations can be executed in place without extra memory movement and overlapped with the original MMA instructions. We demonstrate the efficiency of the fused GEMM operator in \S\ref{sec:matrix}.

With a group size of $128$ and a high precision channel size of $128$, \qq{} has an effective bit of $4.25$\footnote{With $4$-bit for normal values, $8$-bit for outliers, and $16$-bit scale per group, the effective bit is calculated as $((4096-128)*4+128*8)/4096+16/128=4.25$.} on \lm{}-7b. The \textit{effective bit} is defined as the average bits used for each element, including the quantization parameters. This metric is widely used in previous works on \woq{}~\cite{gptq,awq}, mainly because it represents the actual compression ratio and, therefore, the speedup in the memory-bound setting. However, the main benefit of \waq{} in serving scenarios is the computation efficiency of leveraging low-bit arithmetic units instead of the memory reduction. Therefore, we will not use this metric in the following discussions.

\subsection{Dynamic quantization process}
Although fine-grained quantization can better preserve the local variations inside each channel of activations, this advantage would diminish if we statically calculated the quantization parameters based on calibration data, as the actual input might have a different local distribution.

Therefore, \qq{} adopts \textit{dynamic quantization}, tailoring quantization parameters for each activation matrix during inference. To tame the overhead of dynamic quantization, we fuse quantization operations into the prior operator, akin to the implementation of ZeroQuant~\cite{yao2022zeroquant}. Since the additional operator is element-wise (with a reduction and an element-wise division), the run time of the fused operator is still negligible compared to the time-consuming dense and self-attention layers, as Figure \ref{fig:breakdown} shows.

However, asymmetric quantization can lead to significant run-time overhead due to considerable additional computation (as discussed in \S\ref{sec:background}). To strike a balance between throughput and accuracy, we choose symmetric quantization with a carefully chosen clip threshold. We also incorporate GPTQ~\cite{gptq} when quantizing the weight matrix since this is purely an offline process and offers an accuracy boost without sacrificing runtime efficiency.
\vspace{-7pt}

\begin{figure}[tb!]
\centering
\includegraphics[width=0.8\columnwidth]{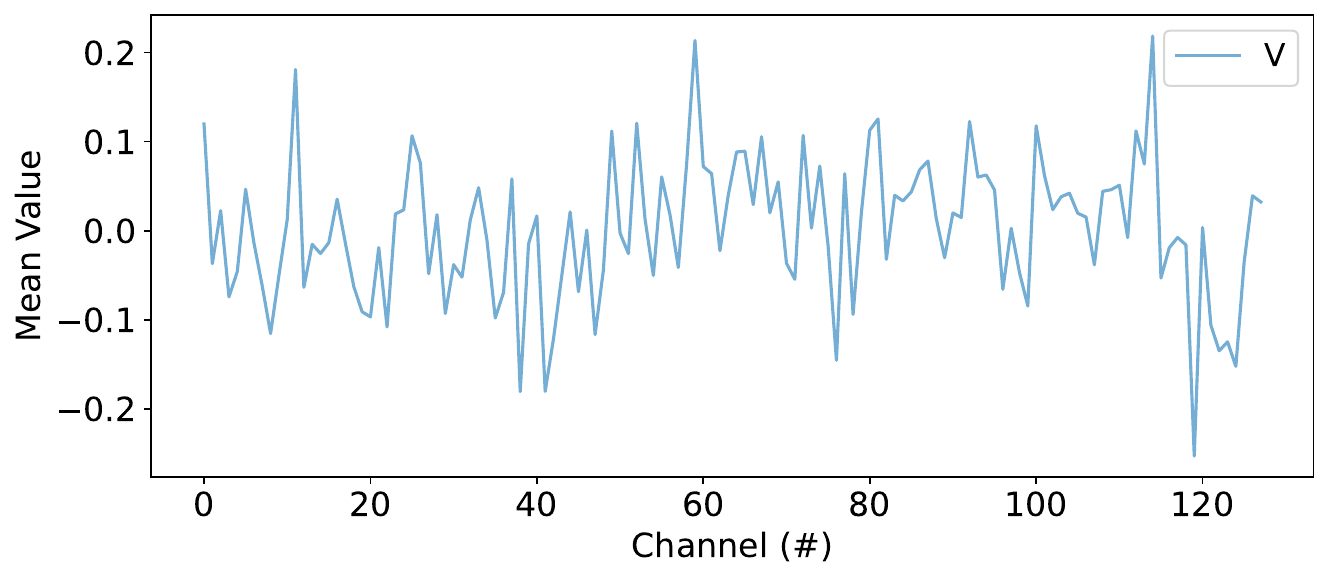}
\vspace{-5pt}
\caption{Sampled value of the V cache within a single attention head from \lm{}-7b. Compared with sampled activations shown in Figure~\ref{fig:design-outlier}, the V cache shows a much less dynamic range with fewer outlier channels, which is much easier to quantize.}
\label{fig:kvcache}
\vspace{-0.25in}
\end{figure}

\subsection{\kv{} quantization}
\label{sec:design-kv_cache}
As described in \S\ref{sec:analyze}, the self-attention layer in the decode stage is highly memory-bound. To mitigate this issue, \qq{} also applies low-bit quantization to the \kv{}. \qq{} loads the \kv{} in low-bit precision and directly dequantizes it before performing the FP16 calculation, which significantly boosts the throughput by large memory reduction. On the other hand, since the memory movement of asymmetric and symmetric quantized \kv{} are similar, they perform similarly on memory-bound self-attention layers. Therefore, \qq{} uses asymmetric quantization on \kv{} as it can provide accuracy benefits. 

Compared with activation matrices, we argue that the \kv{} is more amenable to quantization. To perform self-attention, the Query vector of the incoming token is multiplied by the K cache. The result is normalized using Softmax and further multiplied with the V cache to obtain the output~\cite{vaswani2023attention}. Due to the normalization of Softmax, the quantization error of the K cache has less influence on the output. Furthermore, our profiling in Figure~\ref{fig:kvcache} indicates that the V cache exhibits the outlier phenomenon less frequently, rendering it more suitable for quantization. Therefore, \qq{} directly applies asymmetric low-bit quantization with the granularity of attention head and preserves high accuracy as shown in \S\ref{sec:ablation}.

\subsection{Implementation of quantization workflow}
To demonstrate the feasibility of our design choices, we implement \qq{} on \lm{} models~\cite{touvron2023llama}, as shown in Figure~\ref{fig:workflow}. To leverage the benefit of quantization, \qq{} manages the overhead of the additional operators by kernel fusion: \qq{} fuses quantization operators, including reordering, quantization, and dequantization, into existing operators. For the compute-bound dense layer, \qq{} utilizes the low-bit units to boost throughput. For the memory-bound self-attention layer, \qq{} fuses dequantization with a kernel library for LLM serving, FlashInfer~\cite{flashinfer}, so that only low-bit values from \kv{} are loaded. \qq{} also incorporates PageAttention~\cite{vllm} for efficient memory usage to enable large batch sizes.
\section{Evaluation}
\label{sec:evaluation}

We conduct a comprehensive evaluation of \qq{}'s accuracy and efficiency. For accuracy, we evaluate \qq{} on widely used metrics, generation perplexity and zero-shot accuracy. For efficiency, we evaluate \qq{} from the bottom up, starting with per-kernel performance, followed by end-to-end throughput and latency. We also perform ablation studies to understand how different techniques affect \qq{}, which pinpoints the trade-off between the efficiency and accuracy of each design choice.

\subsection{Quantization setup} 
\qq{} uses symmetric quantization on weights and activations while using asymmetric quantization on the \kv{}. We evaluate \qq{} using a group size of $128$.
 To identify outlier channels, we use $128$ randomly sampled sentences from WikiText2~\cite{wikitext2} as calibration data, following prior works~\cite{owq,shao2023omniquant,liu2023qllm}. We select $128$ channels with the highest square sum values as outlier channels and keep them in INT8. We then reorder activation and weight matrices according to the indices of outlier channels. After reordering, \qq{} adopts GPTQ~\cite{gptq} for the quantization on weight matrices. For clipping, we use a grid search to find optimal clipping factors $0.9$ and $0.85$ for activation and weight quantization, respectively.

\begin{table*}[]
\centering
\caption{
    \textbf{Zero-shot accuracy} of quantized Llama models on six common sense tasks. %
}
\vspace{2pt}
\resizebox{0.91\textwidth}{!}{
\begin{tabular}{cclccccccc}
\toprule
                       &                          &                              & \multicolumn{7}{c}{Zero-shot Accuracy $\uparrow$}                                                                                                                                                                                                                                                       \\ \cline{4-10} 
\multirow{-2}{*}{Size} & \multirow{-2}{*}{\#Bits} & \multirow{-2}{*}{Method}     & PIQA                                   & ARC-e                                  & ARC-c                                  & BoolQ                                  & HellaSwag                              & Winogrande                             & Avg.                                   \\ \midrule
                       & FP16                     & -                            & 77.37                                  & 52.53                                  & 41.38                                  & 73.12                                  & 72.99                                  & 66.85                                  & 64.04                                  \\ \cline{2-10} 
                       &                          & SmoothQuant                  & 63.11                                  & 40.03                                  & 31.57                                  & 58.47                                  & 43.38                                  & 52.80                                  & 48.23                                  \\
                       &                          & OmniQuant                    & 66.15                                  & 45.20                                  & 31.14                                  & 63.51                                  & 56.44                                  & 53.43                                  & 52.65                                  \\
                       &                          & QLLM                         & 68.77                                  & 45.20                                  & 31.14                                  & -                                      & 57.43                                  & 56.67                                  & 51.84                                  \\
                       & \multirow{-4}{*}{W4A4}   & \cellcolor[HTML]{EFEFEF}Atom & \cellcolor[HTML]{EFEFEF}\textbf{76.28} & \cellcolor[HTML]{EFEFEF}\textbf{52.10} & \cellcolor[HTML]{EFEFEF}\textbf{38.99} & \cellcolor[HTML]{EFEFEF}\textbf{69.79} & \cellcolor[HTML]{EFEFEF}\textbf{69.81} & \cellcolor[HTML]{EFEFEF}\textbf{63.69} & \cellcolor[HTML]{EFEFEF}\textbf{61.78} \\ \cline{2-10} 
                       &                          & SmoothQuant                  & 48.69                                  & 25.97                                  & 28.16                                  & 45.26                                  & 26.02                                  & 49.57                                  & 37.28                                  \\
                       &                          & OmniQuant                    & 49.78                                  & 27.19                                  & 27.22                                  & 37.86                                  & 25.64                                  & 49.96                                  & 36.28                                  \\
\multirow{-8}{*}{7B}   & \multirow{-3}{*}{W3A3}   & \cellcolor[HTML]{EFEFEF}Atom & \cellcolor[HTML]{EFEFEF}65.56          & \cellcolor[HTML]{EFEFEF}41.41          & \cellcolor[HTML]{EFEFEF}30.72          & \cellcolor[HTML]{EFEFEF}61.77          & \cellcolor[HTML]{EFEFEF}53.19          & \cellcolor[HTML]{EFEFEF}55.56          & \cellcolor[HTML]{EFEFEF}51.37          \\ \midrule
                       & FP16                     & -                            & 79.05                                  & 59.85                                  & 44.62                                  & 68.53                                   & 76.22                                  & 70.09                                  & 66.39                                  \\ \cline{2-10}
                       &                          & SmoothQuant                  & 64.47                                  & 41.75                                  & 30.89                                  & 62.29                                  & 46.68                                  & 51.70                                  & 49.63                                  \\
                       &                          & OmniQuant                    & 69.69                                  & 47.39                                  & 33.10                                  & 62.84                                  & 58.96                                  & 55.80                                  & 54.63                                  \\
                       &                          & QLLM                         & 71.38                                  & 47.60                                  & 34.30                                  & -                                      & 63.70                                  & 59.43                                  & 55.28                                  \\
                       & \multirow{-4}{*}{W4A4}   & \cellcolor[HTML]{EFEFEF}Atom & \cellcolor[HTML]{EFEFEF}\textbf{77.69} & \cellcolor[HTML]{EFEFEF}\textbf{57.58} & \cellcolor[HTML]{EFEFEF}\textbf{42.92} & \cellcolor[HTML]{EFEFEF}\textbf{67.46} & \cellcolor[HTML]{EFEFEF}\textbf{73.77} & \cellcolor[HTML]{EFEFEF}\textbf{68.51} & \cellcolor[HTML]{EFEFEF}\textbf{64.66} \\ \cline{2-10}
                       &                          & SmoothQuant                  & 47.99                                  & 26.30                                  & 27.65                                  & 46.91                                  & 25.65                                  & 49.64                                  & 37.36                                  \\
                       &                          & OmniQuant                    & 50.22                                  & 26.77                                  & 27.82                                  & 37.83                                  & 25.77                                  & 51.07                                  & 36.58                                  \\
\multirow{-8}{*}{13B}  & \multirow{-3}{*}{W3A3}   & \cellcolor[HTML]{EFEFEF}Atom & \cellcolor[HTML]{EFEFEF}70.08          & \cellcolor[HTML]{EFEFEF}47.94          & \cellcolor[HTML]{EFEFEF}33.70          & \cellcolor[HTML]{EFEFEF}63.46          & \cellcolor[HTML]{EFEFEF}62.93          & \cellcolor[HTML]{EFEFEF}56.75          & \cellcolor[HTML]{EFEFEF}55.81          \\ \midrule
                       & FP16                     & -                            & 80.20                                  & 58.92                                  & 45.31                                  & 68.38                                  & 79.23                                  & 72.69                                  & 67.46                                  \\ \cline{2-10} 
                       &                          & SmoothQuant                  & 59.30                                  & 36.74                                  & 28.58                                  & 59.97                                  & 34.84                                  & 49.96                                  & 44.90                                  \\
                       &                          & OmniQuant                    & 71.21                                  & 49.45                                  & 34.47                                  & 65.33                                  & 64.65                                  & 59.19                                  & 57.38                                  \\
                       &                          & QLLM                         & 73.83                                  & 50.67                                  & 38.40                                  & -                                      & 67.91                                  & 58.56                                  & 57.87                                  \\
                       & \multirow{-4}{*}{W4A4}   & \cellcolor[HTML]{EFEFEF}Atom & \cellcolor[HTML]{EFEFEF}\textbf{78.73} & \cellcolor[HTML]{EFEFEF}\textbf{58.92} & \cellcolor[HTML]{EFEFEF}\textbf{45.82} & \cellcolor[HTML]{EFEFEF}\textbf{68.47} & \cellcolor[HTML]{EFEFEF}\textbf{77.40} & \cellcolor[HTML]{EFEFEF}\textbf{73.09} & \cellcolor[HTML]{EFEFEF}\textbf{67.07} \\ \cline{2-10} 
                       &                          & SmoothQuant                  & 49.46                                  & 27.53                                  & 28.16                                  & 39.42                                  & 26.05                                  & 51.38                                  & 37.00                                  \\
\multirow{-7}{*}{30B}  & \multirow{-2}{*}{W3A3}   & \cellcolor[HTML]{EFEFEF}Atom & \cellcolor[HTML]{EFEFEF}72.47          & \cellcolor[HTML]{EFEFEF}49.54          & \cellcolor[HTML]{EFEFEF}37.80          & \cellcolor[HTML]{EFEFEF}65.75          & \cellcolor[HTML]{EFEFEF}66.99          & \cellcolor[HTML]{EFEFEF}60.14          & \cellcolor[HTML]{EFEFEF}58.78          \\ \midrule
                       & FP16                     & -                            & 80.79                                  & 58.71                                  & 46.33                                  & 82.26                                  & 80.71                                  & 77.03                                  & 70.97                                  \\ \cline{2-10} 
                       &                          & SmoothQuant                  & 60.72                                  & 38.80                                  & 30.29                                  & 57.61                                  & 36.81                                  & 53.43                                  & 46.28                                  \\
                       &                          & OmniQuant                    & 71.81                                  & 48.02                                  & 35.92                                  & 73.27                                  & 66.81                                  & 59.51                                  & 59.22                                  \\
                       &                          & QLLM                         & 73.56                                  & 52.06                                  & 39.68                                  & -                                      & 70.94                                  & 62.90                                  & 59.83                                  \\
                       & \multirow{-4}{*}{W4A4}   & \cellcolor[HTML]{EFEFEF}Atom & \cellcolor[HTML]{EFEFEF}\textbf{80.41} & \cellcolor[HTML]{EFEFEF}\textbf{58.12} & \cellcolor[HTML]{EFEFEF}\textbf{45.22} & \cellcolor[HTML]{EFEFEF}\textbf{82.02} & \cellcolor[HTML]{EFEFEF}\textbf{79.10} & \cellcolor[HTML]{EFEFEF}\textbf{72.53} & \cellcolor[HTML]{EFEFEF}\textbf{69.57} \\ \cline{2-10} 
                       &                          & SmoothQuant                  & 49.56                                  & 26.64                                  & 29.10                                  & 42.97                                  & 26.05                                  & 51.14                                  & 37.58                                  \\
\multirow{-7}{*}{65B}  & \multirow{-2}{*}{W3A3}   & \cellcolor[HTML]{EFEFEF}Atom & \cellcolor[HTML]{EFEFEF}75.84          & \cellcolor[HTML]{EFEFEF}51.43          & \cellcolor[HTML]{EFEFEF}41.30          & \cellcolor[HTML]{EFEFEF}74.07          & \cellcolor[HTML]{EFEFEF}72.22          & \cellcolor[HTML]{EFEFEF}64.33          & \cellcolor[HTML]{EFEFEF}63.20          \\ \bottomrule
\end{tabular}
}
\label{table:llama_zeroshot}
\end{table*}
\begin{table*}[]
\centering
\caption{
    \textbf{Perplexity} of quantized Llama models on WikiText2, PTB and C4 dataset. %
}
\vspace{2pt}
\resizebox{0.95\textwidth}{!}{
\begin{tabular}{cclccc}
\toprule
                       &                        &                              & \multicolumn{3}{c}{Perplexity $\downarrow$}                                                                                        \\
\multirow{-2}{*}{Size} & \multirow{-2}{*}{Bits} & \multirow{-2}{*}{Method}     & WikiText2                             & PTB                                   & C4                                    \\ \midrule
                       & FP16                   & -                            & 5.68                                  & 8.80                                  & 7.08                                  \\ \cline{2-6} 
                       &                        & SmoothQuant                  & 22.62                                 & 40.69                                 & 31.21                                 \\
                       &                        & OmniQuant                    & 11.59                                 & 20.65                                 & 14.96                                 \\
                       &                        & QLLM                         & 9.65                                  & -                                     & 12.29                                 \\
                       & \multirow{-4}{*}{W4A4} & \cellcolor[HTML]{EFEFEF}Atom & \cellcolor[HTML]{EFEFEF}\textbf{6.16} & \cellcolor[HTML]{EFEFEF}\textbf{9.62} & \cellcolor[HTML]{EFEFEF}\textbf{7.70} \\ \cline{2-6} 
                       &                        & SmoothQuant                  & 2.7e4                                 & 3.5e4                                 & 2.6e4                                 \\
                       &                        & OmniQuant                    & 3.4e3                                 & 7.5e3                                 & 6.3e3                                 \\
\multirow{-8}{*}{7B}   & \multirow{-3}{*}{W3A3} & \cellcolor[HTML]{EFEFEF}Atom & \cellcolor[HTML]{EFEFEF}11.77         & \cellcolor[HTML]{EFEFEF}20.84         & \cellcolor[HTML]{EFEFEF}15.43         \\ \midrule
                       & FP16                   & -                            & 4.10                                  & 7.30                                  & 5.98                                  \\ \cline{2-6} 
                       &                        & SmoothQuant                  & 109.85                                & 142.34                                & 87.06                                 \\
                       &                        & OmniQuant                    & 10.34                                 & 14.91                                 & 12.49                                 \\
                       &                        & QLLM                         & 8.37                                  & -                                     & 11.51                                 \\
                       & \multirow{-4}{*}{W4A4} & \cellcolor[HTML]{EFEFEF}Atom & \cellcolor[HTML]{EFEFEF}\textbf{4.54} & \cellcolor[HTML]{EFEFEF}\textbf{7.69} & \cellcolor[HTML]{EFEFEF}\textbf{6.35} \\ \cline{2-6} 
                       &                        & SmoothQuant                  & 1.5e4                                 & 1.6e4                                 & 1.5e4                                 \\
\multirow{-7}{*}{30B}  & \multirow{-2}{*}{W3A3} & \cellcolor[HTML]{EFEFEF}Atom & \cellcolor[HTML]{EFEFEF}6.94          & \cellcolor[HTML]{EFEFEF}12.12         & \cellcolor[HTML]{EFEFEF}9.14          \\ \bottomrule
\end{tabular}
\begin{tabular}{cclccc}
\toprule
                       &                        &                              & \multicolumn{3}{c}{Perplexity $\downarrow$}                                                                                        \\
\multirow{-2}{*}{Size} & \multirow{-2}{*}{Bits} & \multirow{-2}{*}{Method}     & WikiText2                             & PTB                                   & C4                                    \\ \midrule
                       & FP16                   & -                            & 5.09                                  & 8.07                                  & 6.61                                  \\ \cline{2-6} 
                       &                        & SmoothQuant                  & 33.98                                 & 73.83                                 & 41.53                                 \\
                       &                        & OmniQuant                    & 10.90                                 & 18.03                                 & 13.78                                 \\
                       &                        & QLLM                         & 8.41                                  & -                                     & 10.58                                 \\
                       & \multirow{-4}{*}{W4A4} & \cellcolor[HTML]{EFEFEF}Atom & \cellcolor[HTML]{EFEFEF}\textbf{5.46} & \cellcolor[HTML]{EFEFEF}\textbf{8.60} & \cellcolor[HTML]{EFEFEF}\textbf{7.03} \\ \cline{2-6} 
                       &                        & SmoothQuant                  & 1.3e4                                 & 1.6e4                                 & 1.5e4                                 \\
                       &                        & OmniQuant                    & 7.2e3                                 & 1.6e4                                 & 1.3e4                                 \\
\multirow{-8}{*}{13B}  & \multirow{-3}{*}{W3A3} & \cellcolor[HTML]{EFEFEF}Atom & \cellcolor[HTML]{EFEFEF}8.40          & \cellcolor[HTML]{EFEFEF}15.84         & \cellcolor[HTML]{EFEFEF}10.81         \\ \midrule
                       & FP16                   & -                            & 3.53                                  & 6.91                                  & 5.62                                  \\ \cline{2-6} 
                       &                        & SmoothQuant                  & 88.89                                 & 278.76                                & 283.80                                \\
                       &                        & OmniQuant                    & 9.18                                  & 16.18                                 & 11.31                                 \\
                       &                        & QLLM                         & 6.87                                  & -                                     & 8.98                                  \\
                       & \multirow{-4}{*}{W4A4} & \cellcolor[HTML]{EFEFEF}Atom & \cellcolor[HTML]{EFEFEF}\textbf{3.89} & \cellcolor[HTML]{EFEFEF}\textbf{7.22} & \cellcolor[HTML]{EFEFEF}\textbf{5.92} \\ \cline{2-6} 
                       &                        & SmoothQuant                  & 6.6e8                                 & 3.7e8                                 & 4.4e8                                 \\
\multirow{-7}{*}{65B}  & \multirow{-2}{*}{W3A3} & \cellcolor[HTML]{EFEFEF}Atom & \cellcolor[HTML]{EFEFEF}5.89          & \cellcolor[HTML]{EFEFEF}9.71          & \cellcolor[HTML]{EFEFEF}7.94          \\ \bottomrule
\end{tabular}
}
\label{table:llama_ppl}
\end{table*}

For the preprocessing of weight quantization and outlier identification, we run \qq{} on a single RTX Ada 6000 and quantize the model layer-by-layer. For large \lm{}-65B, \qq{} takes roughly $4$ hours to complete the process.

\subsection{Accuracy evaluation}
\textbf{Benchmarks.}
We evaluate \qq{} on popular open-sourced \lm{}~\cite{touvron2023llama} models. We focus on low-bit settings, INT4 and INT3 \waq{}. We adopt commonly used metrics of model accuracy, perplexity, and zero-shot accuracy. For perplexity, we evaluate on WikiText2~\cite{wikitext2}, PTB~\cite{PTB}, and C4~\cite{c4} datasets. For zero-shot tasks, we use lm-eval~\cite{eval-harness}, based on which we evaluate \qq{} on PIQA~\cite{bisk2019piqa}, ARC \cite{arc}, BoolQ \cite{clark2019boolq}, HellaSwag \cite{zellers2019hellaswag}, and WinoGrande \cite{sakaguchi2019winogrande} tasks.

\textbf{Baselines.}
We compare \qq{} to recently released post-training quantization techniques: SmoothQuant~\cite{smoothquant}, OmniQuant~\cite{shao2023omniquant}, and QLLM~\cite{liu2023qllm}. For SmoothQuant, we implement our own version as the official code does not support \lm{} models and only has W8A8 quantization. We conducted a grid search on the $alpha$ value defined in SmoothQuant and reported the best numbers for each benchmark. For OmniQuant, we use their pre-quantized weights for W4A4 evaluations and evaluate W3A3 by running their official code. To obtain the best W3A3 results for OmniQuant, we conduct a hyperparameter search and identify $lr = 1e^{-4}$ and $alpha = 0.75$ for their quantization process. We skip W3A3 OmniQuant on \lm{}-30B and \lm{}-65B due to the large resource requirement of its quantization process. For QLLM, we report the W4A4 numbers in their paper but do not evaluate W3A3 as their code was unavailable when we conducted experiments.

\textbf{Zero-shot accuracy.}
Table~\ref{table:llama_zeroshot} compares the zero-shot accuracy of six tasks between \qq{} and baselines on \lm{} models. \qq{} significantly outperforms the other \waq{} methods. For W4A4, \qq{} shows only a $2.3\%$, $1.7\%$, $0.4\%$ and $1.4\%$ average accuracy loss for Llama at $7$B, $13$B, $30$B and $65$B sizes when compared to FP16. At the same time, previous works showed a $9.6\%$ to $23.8\%$ accuracy loss under the same settings.

\textbf{Perplexity.}
Table~\ref{table:llama_ppl} reports perplexity results of \qq{} and baselines on \lm{} models. As the table shows, though recent methods such as OmniQuant and QLLM successfully reduce the perplexity of W4A4 to around $10$, the accuracy loss is still significant. \qq{} further reduces the perplexity and achieves less than $0.4$ perplexity increase on all three datasets with \lm{}-65b. For W3A3, \qq{} still largely maintains the perplexity, with an average $2.3$ perplexity increase for \lm{}-65B. At the same time, existing works do not achieve acceptable perplexity. Note that \qq{} has less accuracy loss when quantizing larger models.

\begin{figure*}[t!]
    \centering
    \subfigure[End-to-end throughput]{
        \includegraphics[width=0.28\linewidth]{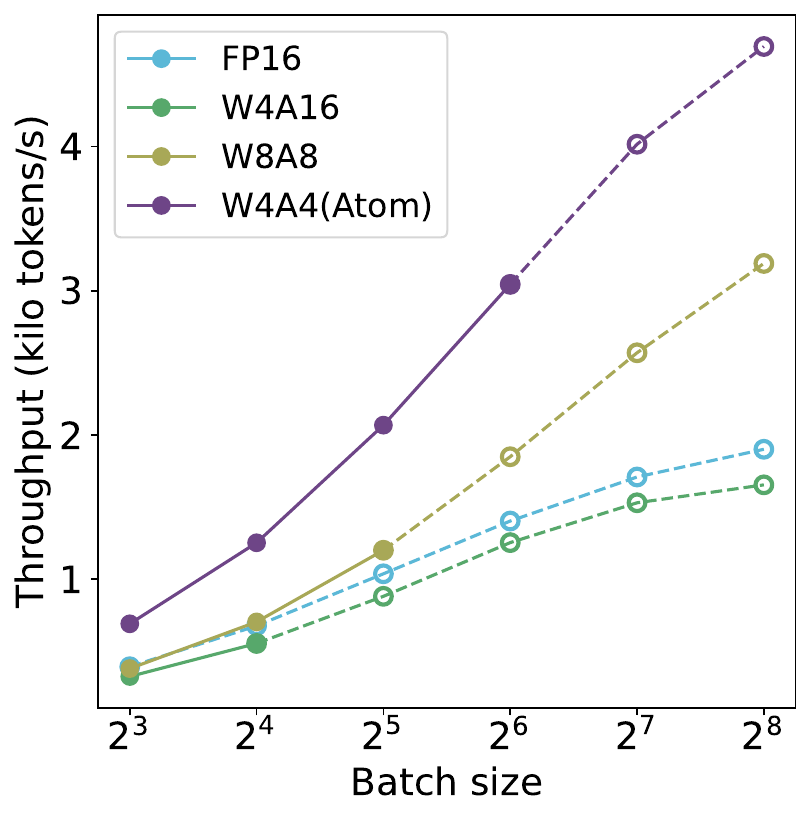}
        \label{fig:e2e-throughput}
    }
    \subfigure[Decode latency per token]{
        \includegraphics[width=0.295\linewidth]{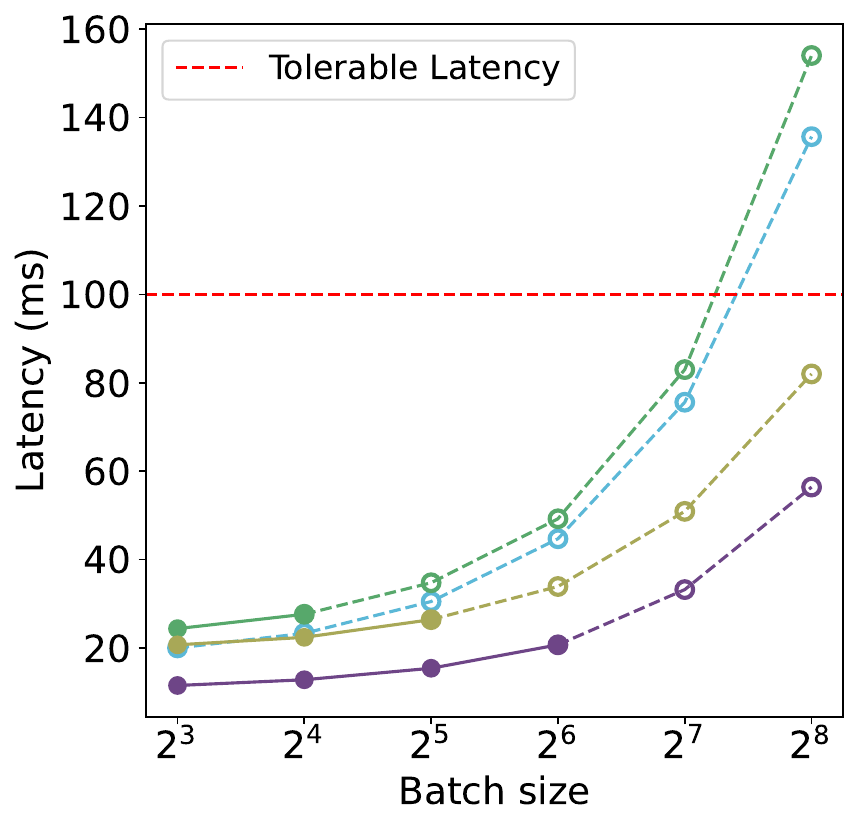}
        \label{fig:e2e-latency}
    }
    \subfigure[Comparison given fixed GPU memory]{
        \includegraphics[width=0.322\linewidth]{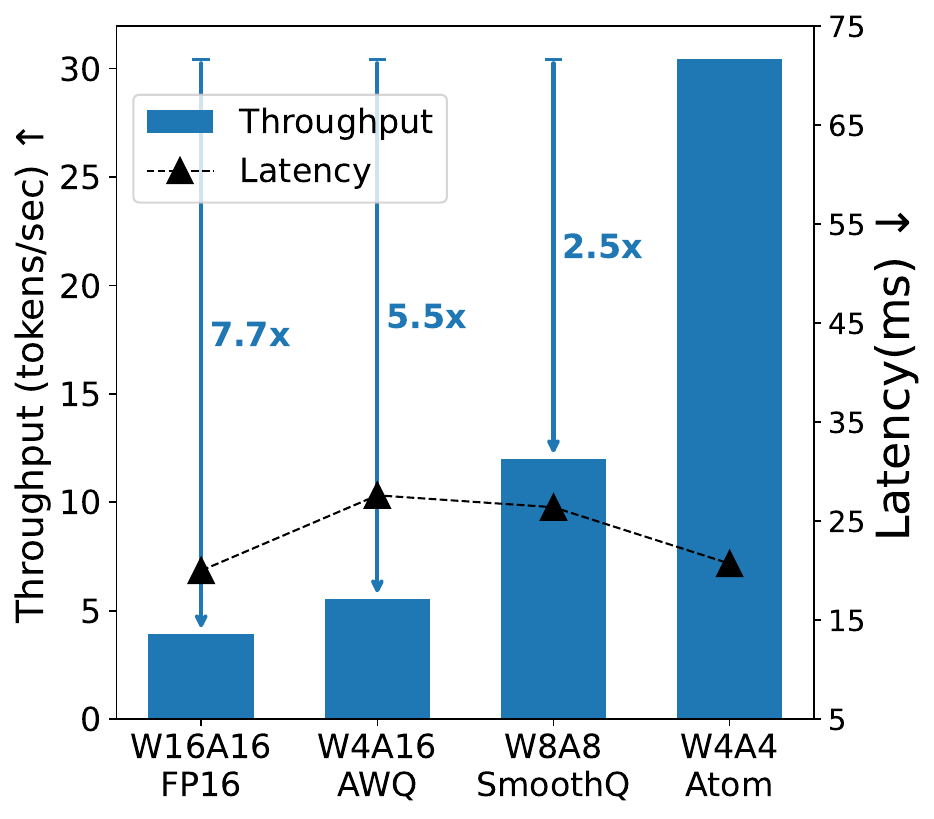}
        \label{fig:e2e-fix}
    }
    \caption{End-to-end evaluation of \qq{}. Solid lines are exact measurements, while dashed lines are estimations due to the limited memory capacity. (a) The number of generated tokens per second. (b) Average decode latency per token. \qq{} surpasses all other quantization methods for both throughput and latency. (c) Performance evaluated under a fixed amount of GPU memory. Note that \qq{} boosts the throughput by $2.5\times$ more than W8A8 since it enables a larger batch size, which utilizes the batching effect.}
    \label{fig:e2e-all}
    \vspace{-0.15in}
\end{figure*}

\subsection{Efficiency evaluation}
To demonstrate the efficiency of \qq{}, we conduct experiments profiling both per-kernel and end-to-end performance. Since the highly efficient INT4 arithmetic is supported by NVIDIA GPUs, we evaluate \qq{} with W4A4 quantization on a 24GB RTX 4090 with CUDA 11.3.

\subsubsection{Kernel evaluation}
\label{sec:matrix}
\begin{figure}[tb!]
    \centering
    \subfigure[Dense layer]{
        \includegraphics[width=0.45\columnwidth]{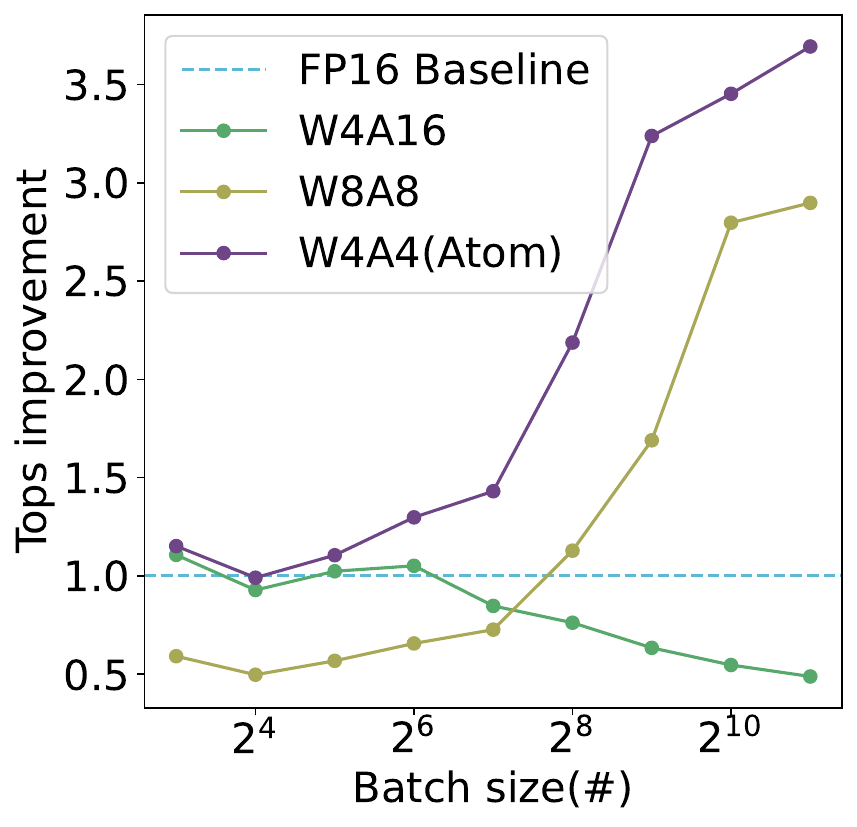}
        \label{fig:batch-dense}
    }
    \subfigure[Self-attention]{
        \includegraphics[width=0.47\columnwidth]{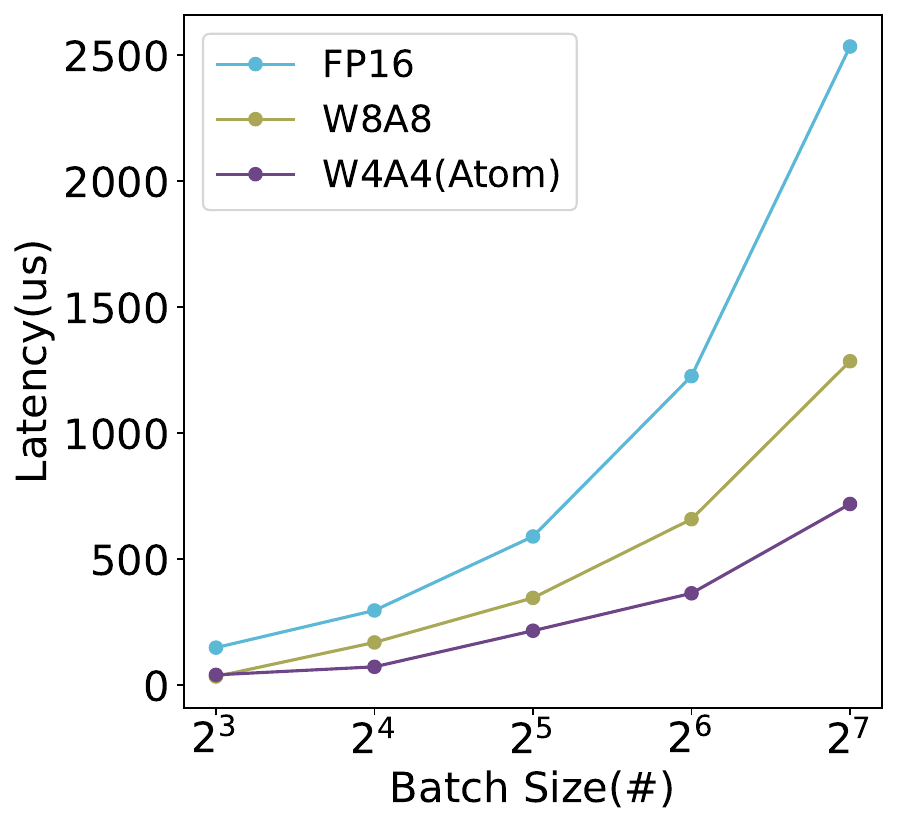}
        \label{fig:batch-self-attn}
    }
    \caption{Performance evaluation of different quantization approaches on \qq{} and baseline kernels. We set up the evaluation configuration aligned with the \lm{}-7b config and $1024$ sequence length. Kernels are evaluated by NVBench~\cite{nvidia_nvbench}.}
    \label{fig:batch-evaluate}
    \vspace{-0.2in}
\end{figure}

\textbf{Matrix multiplication.}
We evaluate the fused GEMM operator implemented by \qq{}, as shown in Figure~\ref{fig:batch-dense}. We also implemented fused GEMM for 8-bit \waq{} (W8A8) and 4-bit \woq{} (W4A16) following the existing work~\cite{smoothquant,awq} as baselines. For smaller batch sizes, GEMM is memory-bound; thus, \woq{}'s memory reduction is effective. However, as the batch size increases, the efficiency of \woq{} diminishes in the compute-bound setting due to the expensive FP16 calculations. At the same time, 4-bit \qq{} outperforms all other approaches due to its hardware efficiency. At batch size $512$, \qq{}'s matrix-multiplication achieves $3.4\times$ and $1.9\times$ speedup over FP16 and INT8 kernels.

\textbf{Self-attention.}
For the self-attention layer, we fuse different quantization methods into FlashInfer~\cite{flashinfer}, which is a performant kernel library for LLMs serving. We also integrate PageAttention~\cite{vllm} for efficient memory usage. We evaluate our implementation and show the results in Figure~\ref{fig:batch-self-attn}. The decrease in bits linearly reduces the memory usage of the \kv{}, therefore proportionally boosting the throughput in the memory-bound setting. At batch size $128$, \qq{} achieves a $1.8\times$ speedup over INT8 quantization and $3.5\times$ over the FP16 baseline.

\subsubsection{End-to-end evaluation}
\textbf{Serving setup.}
We integrate \qq{} into Punica, an LLM serving framework~\cite{punica}, to evaluate the performance in the end-to-end scenario. We also integrate W8A8 and W4A16 quantizations following previous works~\cite{smoothquant,awq} as baselines. To generate a representative workload, we use ShareGPT~\cite{ShareGPT} to collect the distribution of prefill and decode request length. We treat multi-round conversations as requests from multiple users. Specifically, we concatenate all previous prompts and responses and use them as the prompt for the new user request. We vary the batch size from $8$ to $256$, which represents the practical range in LLM serving\footnote{With quantization, pipelining, and tensor parallelism to amortize weights, it is practical to deploy a $180$B model with a $256$ batch size in the serving scenario~\cite{splitwise}.}. All requests are served in a First-Come-First-Served manner. When a request is finished, we re-fill the on-the-fly batch with a new request following \textit{continous batching} as introduced in Orca~\cite{orca}. Due to GPU memory limits, we only show the exact results on small batch sizes. When the memory requirement cannot be satisfied, we simulate the performance by reusing the KV-caches from a smaller batch size while preserving the data access pattern and amount of computation.

\textbf{End-to-end throughput.}
We show the end-to-end throughput, i.e., generated tokens per second, in Figure~\ref{fig:e2e-throughput}. Solid lines represent exact evaluation results, while dashed lines represent our simulated results for the cases that exceed our GPU's memory capacity. As Figure~\ref{fig:e2e-throughput} shows, \qq{} outperforms other quantization methods on all batch sizes. If we fix the available memory as shown in Figure~\ref{fig:e2e-fix}, \qq{} can achieve larger batch sizes so that its throughput further surpasses all baselines while still meeting the latency target. \qq{} achieves \etespeedupfp{} throughput compared to the FP16 baseline and \etespeedupsq{} throughput compared to INT8 quantization using the same amount of memory. In contrast, \woq{} is bounded by FP16 computation capacity in dense layers and large memory movement of the \kv{} in the self-attention layer.

\textbf{End-to-end latency.}
We measure the latency as the average decoding time of each token, without considering the queuing time. \qq{} significantly outperforms other quantization methods on every batch size. When we achieve the highest practical performance at batch size $64$, our latency is lower than INT8 or FP16 implementations, even under batch size $8$. Notably, even at batch size $256$, our latency is still lower than $100$ ms, which has been shown to be the effective reading speed of human eyes by a prior study ~\cite{readingspeed}.

\subsection{Ablation study of quantization techniques}
\label{sec:ablation}
In this subsection, we comprehensively evaluate the effectiveness of quantization techniques used in \qq{}, in terms of both accuracy and efficiency, to better illustrate our design choices and the trade-off between accuracy and efficiency.

\subsubsection{Ablation study to evaluate accuracy}
We examine the accuracy gain or loss of different quantization techniques used in \qq{}. We first use RTN and adopt per-channel quantization for weights and per-token quantization for activations, which is the standard quantization recipe~\cite{smoothquant}, to quantize the model to W4A4. We then apply other quantization techniques used in \qq{}, i.e., mixed-precision, quantizing outliers, group quantization, clipping, GPTQ, and \kv{} quantization, and examine the perplexity case by case. As shown in Table~\ref{table:quant_ablation}, keeping outlier channels in FP16 significantly reduces the perplexity. Further quantizing outliers into INT8 only results in a very minor $0.05$ perplexity increase, which indicates mixed precision effectively addresses the outlier issue. Besides, fine-grained group quantization brings another major perplexity reduction. Furthermore, using clipping and GPTQ lowers perplexity by $0.09$ each. After all, quantizing \kv{} results in a slight $0.12$ perplexity increase, which echoes our finding in Section \ref{sec:design-kv_cache}.

\begin{table}[]
\centering
\caption{
    Ablation study on different quantization techniques used in \qq{}. The model used in this table is \lm{}-7B.
}
\vspace{5pt}
\resizebox{0.8\columnwidth}{!}{
\begin{tabular}{ll}
\toprule
\textbf{Quantization method} & \textbf{WikiText2 PPL$\downarrow$} \\
\midrule
FP16 baseline                & 5.68                   \\ 
\midrule
 W4A4 RTN      & 2315.52                \\ %
+ Keeping 128 outliers in FP16  & 11.34 (2304.2$\downarrow$)                   \\
+ Quantizing outliers to INT8  & 11.39 (0.05$\uparrow$)                   \\
+ Group size 128             & 6.22 (5.17$\downarrow$)                      \\
+ Clipping                   & 6.13 (0.09$\downarrow$)                      \\
+ GPTQ                       & 6.04 (0.09$\downarrow$)                      \\
+ Quantizing \kv{} to INT4            & 6.16 (0.12$\uparrow$)                      \\
\bottomrule
\end{tabular}
}
\label{table:quant_ablation}
\vspace{-0.15in}
\end{table}

\vspace{-10pt}

\subsubsection{Ablation study to evaluate efficiency}
\label{sec:efficiency_ablation}
We then showcase the GEMM kernel throughput with different fused quantization techniques\footnote{Kernel performance is profiled by NVBench~\cite{nvidia_nvbench} with the \lm{}-7b config and a batch size of $4096$ on RTX 4090.}. A pure INT4 GEMM implementation without any quantization operation achieves nearly $980$ TOPS. Fusion of mixed precision, which keeps $128$ channel calculations in INT8 Tensor Cores, leads to $8$\% overhead, with $900$ TOPS throughput. Fine-grained group quantization contributes to the major overhead since it deeply affects the compute pipeline. The fusion of group dequantization decreases the performance to $770$ TOPS. However, the fused GEMM kernel still outperforms the theoretical limit of INT8 throughput by nearly $18$\%.

Besides, to demonstrate the efficiency of channel reordering, we also conduct an ablation study on \qq{} and baseline. The baseline is implemented following the previous work~\cite{llmint8}, with matrix decomposition for mixed precision quantization. At the same time, \qq{} fuses quantization operators, including reordering and quantization, into existing operators. We evaluate batch sizes from $16$ to $256$ and measure the inference latency of a layer norm and a GEMM operation. Results show that \qq{} consistently outperforms the baseline from $25$\% to $35$\%.
\section{Discussion}
With innovations of model architectures like Mixture of Experts (MoE)~\cite{jiang2024mixtral,dai2024deepseekmoe}, State Space Models (SSMs)~\cite{gu2022efficiently,gu2023mamba}, and evolvement of hardware accelerators (e.g., NVIDIA Blackwell GPU \cite{blackwell}), it's important that \qq{} can be used for new models and hardware. In this section, we provide evaluations on more LLMs and data formats.

\begin{table}[bt!]
\centering
\caption{
    WikiText2 perplexity for \lm{}-2 and Mixtral.
}
\vspace{2pt}
\resizebox{\columnwidth}{!}{
\begin{tabular}{clcccc}
\toprule
                          & \multicolumn{1}{c}{}                         & \multicolumn{3}{c}{Llama2}                                                                                            & Mixtral                               \\
\multirow{-2}{*}{\# Bits} & \multicolumn{1}{c}{\multirow{-2}{*}{Method}} & 7B                                    & 13B                                   & 70B                                   & 8x7B                                  \\ \hline
FP16                      & -                                            & 5.47                                  & 4.88                                  & 3.32                                  & 3.84                                  \\ \hline
                          & SmoothQuant                                  & 83.12                                 & 35.88                                 & -                                     & -                                     \\
                          & OmniQuant                                    & 14.61                                 & 12.3                                  & -                                     & -                                     \\
                          & \cellcolor[HTML]{EFEFEF}Atom (INT)           & \cellcolor[HTML]{EFEFEF}\textbf{6.03} & \cellcolor[HTML]{EFEFEF}\textbf{5.27} & \cellcolor[HTML]{EFEFEF}\textbf{3.68} & \cellcolor[HTML]{EFEFEF}\textbf{4.41} \\
\multirow{-4}{*}{W4A4}    & \cellcolor[HTML]{EFEFEF}Atom (FP)            & \cellcolor[HTML]{EFEFEF}\textbf{6.14} & \cellcolor[HTML]{EFEFEF}\textbf{5.35} & \cellcolor[HTML]{EFEFEF}\textbf{3.78} & \cellcolor[HTML]{EFEFEF}\textbf{4.50} \\ %
\bottomrule
\end{tabular}
}
\label{table:llama2_ppl}
\vspace{-10pt}
\end{table}
\vspace{-4pt}

\textbf{Generality on models.}
\qq{}'s main techniques to achieve high accuracy are mixed precision for outliers and fine-grained quantization for normal values. We empirically find these are generalizable to newer transformer-based LLMs. In Table~\ref{table:llama2_ppl}, we show the perplexity results of two relatively new LLMs, \lm{}-2 \cite{touvron2023llama2} and Mixtral~\cite{jiang2024mixtral}. To generalize on MoE models, \qq{} only needs to adapt to using different reorder indices for different experts' FFN\footnote{In practice, we find that accuracy is similar when \qq{} share reorder indices across all experts in an MoE layer. Therefore, we use shared indices for efficiency consideration.}. As Table \ref{table:llama2_ppl} shows, \qq{} still outperforms baselines and maintains high accuracy.

\textbf{Generality on data formats.}
With the support for emerging data formats such as FP4 and MX~\cite{liu2023llmfp4,rouhani2023microscaling} on new hardware, we also evaluate the effectiveness of \qq{} in FP4. As shown in Table~\ref{table:llama2_ppl}, \qq{} maintains a similar accuracy to INT4 when quantizing both weights and activations into FP4. We conclude that the representation capability between INT4 and FP4 is similar. Additionally, group quantization with the MX format is supported by NVIDIA Blackwell GPUs. We expect this hardware feature can mitigate the group quantization overhead of \qq{} as described in \S~\ref{sec:efficiency_ablation}.
\section{Related Work}

\textbf{LLM serving.}
Various works have been explored to improve LLM serving throughput. \cite{scaleinference} investigated the batching effect when scaling up LLMs. Orca~\cite{orca} proposed \textit{continuous batching} to improve GPU utilization by refilling the on-the-fly batch. vLLM~\cite{vllm} utilized page tables to manage \kv{}, which significantly increases GPU memory utilization. FlexGen~\cite{flexgen} proposed an offload mechanism to support larger batches for high serving throughput. However, unlike prior works, in this paper, we delve deep into the intersection between quantization and LLM serving. 

\textbf{Weight-only quantization.}
For LLMs, weight matrices lead to large memory movement, limiting decode efficiency. Weight-only quantization uses low-bit precision to approximate weight matrices. For instance, GPTQ~\cite{gptq} used 4-bit to quantize the weight based on the approximate second-order information. AWQ~\cite{awq} further advanced accuracy by preserving salient weights. SqueezeLLM~\cite{kim2023squeezellm} handled outliers through non-uniform quantization and used a sparse format to keep outliers and sensitive weights at high precision. QuiP~\cite{chee2023quip} successfully represented weights using 2-bit by an adaptive rounding method. Nonetheless, in the LLM serving scenario, the overhead of loading the weight matrix is amortized due to batching. Thus, the dense layer becomes compute-bound, while \woq{} fails to use efficient low-bit hardware to deliver ideal throughput.

\textbf{Weight-activation quantization.}
Weight-activation quantization quantizes both the weight and activation matrices, which is considered more challenging due to the outlier phenomenon of the activation. LLM.INT8~\cite{llmint8} proposed mixed precision to preserve outlier values in activation matrices. \cite{smoothquant, shao2023omniquant, yao2022zeroquant, wei2023outlier} used mathematical equivalent transformations to manage activation outliers. RPTQ~\cite{yuan2023rptq} rearranges the channels to reduce the variance within one quantization group, further enhancing the accuracy. Some works~\cite{liu2023qllm, wu2023zeroquantfp} used low-rank matrices to compensate for quantization error. Others~\cite{oliveguo,zhou2023sysmol} used algorithm and architecture co-design to accommodate outliers. However, these approaches either suffer significant accuracy loss at extremely low-bit precision or lack practical hardware support. In this work, our method achieves notable accuracy with low-bit representation and ensures practical speedup.
\section{Conclusion}
We presented \qq{}, a low-bit quantization method that leverages the underlying hardware efficiently to achieve both high accuracy and high throughput for LLM serving. We use mixed-precision quantization with reordering, fine-grained group quantization, dynamic quantization, and \kv{} quantization to preserve accuracy while fully exploiting emerging low-bit hardware support. We integrate \qq{} into an end-to-end serving framework, achieving up to \etespeedupfp{} throughput enhancement compared to the FP16 baseline as well as maintaining less than $1.4$\% zero-shot accuracy loss.
\section*{Acknowledgments}

We thank Jiaming Tang and Yixin Dong for their discussion and insightful feedback. This work was supported in part by ACE and PRISM, two of the seven centers in JUMP 2.0, a Semiconductor Research Corporation (SRC) program sponsored by DARPA; by the National Science Foundation (NSF) under grant CCF-1518703 and award CNS-2211882; and by DARPA under the RTML program. The work was also supported by gifts from Qualcomm and Intel (TSA center).

\bibliography{mlsys}
\bibliographystyle{mlsys2024}

\appendix

\end{document}